\documentclass{article} 
\usepackage{iclr2022_srml_workshop,times}
\usepackage{graphicx}

\usepackage{hyperref}
\usepackage{url}

\title{Towards learning to explain with concept bottleneck models: mitigating information leakage}

\author{Joshua Lockhart\thanks{\texttt{joshualockhart@gmail.com}}\\
J.P. Morgan AI Research\\
London, UK
\And
Nicol\'as Marchesotti \\
J.P. Morgan AI Research\\
London, UK
\And
Daniele Magazzeni \\
J.P. Morgan AI Research\\
London, UK
\AND
Manuela Veloso \\
J.P. Morgan AI Research\\
New York, NY, USA
}

\iclrfinalcopy 
\begin{document}

\maketitle

\begin{abstract}
Concept bottleneck models perform classification by first predicting which of a list of human provided \textit{concepts} are true about a datapoint. Then a downstream model uses these predicted concept labels to predict the target label.
The predicted concepts act as a rationale for the target prediction.
Model trust issues emerge in this paradigm when soft concept labels are used: it has previously been observed that extra information about the data distribution leaks into the concept predictions.
In this work we show how Monte-Carlo Dropout can be used to attain soft concept predictions that do not contain leaked information.
\end{abstract}

Concept bottleneck models \citep{CBMs}, and more broadly, concept-based learning techniques \citep{Kim2018InterpretabilityBF, 2020, kazhdan2020now, 9658985, bahadori2021debiasing, completeness} have emerged as a compelling way of aligning machine learning models with \textit{concepts} that are understandable by human beings. The concept bottleneck model (CBM), which is the focus of this paper, consists of two neural networks: the \textit{concept learning model} (CLM), and the \textit{target classifier}. The CLM model takes an input data point and predicts which of a number of pre-defined binary concepts are true about that datapoint, while the target classifier takes these predicted concepts and uses them to predict the final target label.

The Caltech-UCSD Birds dataset \citep{wah2011caltech} is a natural setting for applying concept bottleneck models. This dataset consists of images of birds, the task is to predict the species of the bird. Each bird image is also annotated with a number of pre-defined binary \textit{concepts}, describing physical attributes of the bird in question such as if it has stripes on its belly, has a pointed beak, \textit{etc}.
The CLM of a concept bottleneck model is trained to predict the binary concepts, then the target model is trained to predict the species of the bird from those predicted concepts. In this way, one can split a complicated classification into a number of smaller subtasks, adding transparency to the prediction process. Furthermore, predictions have a natural interpretation in terms of concepts: `I believe this to be a (\textit{$target$}) \textbf{parrot} because I see (\textit{$Concept_i$}) a \textbf{pointy peak} and (\textit{$Concept_j$}) \textbf{red plumage}'.

However, issues emerge in this paradigm that can erode trust that the components of the CBM are behaving as intended.
\citet{margeloiu2021concept} question whether the CLM concept predictions are based on the correct parts of the input data. \citet{mahinpei2021promises} show that information about the input data can leak into the concept predictions, causing the target classifier to perform absurdly well even when it shouldn't. In this work we tackle the latter issue of information leakage. Leakage occurs when soft concept predictions are used. We show how Monte-Carlo Dropout \citep{MCD} can be employed to produce soft concept predictions that do not leak information in this way.

\section{Concept bottleneck models and information leakage}
Concept bottleneck models 
\citep{CBMs}
consider classification problems where each data point $X$ has, in addition to a ground truth class label $Y$, a list of ground truth binary concept labels $C_1,\dots, C_L$. These concepts encode information about the data point $X$ that could be useful for predicting the ground truth label $Y$. Returning to the example given in the previous section, the bird species classification problem would have the species of the bird provided as the ground truth label $Y$, and the bird's attributes encoded as the binary concept labels (\textit{e.g.,} $C_1$ is true if plumage is red, $C_2$ is true if plumage is brown, $C_{31}$ is true if beak is pointed).

Concept bottleneck models attack such classification problems by training a \textit{concept labeling model} (CLM) $f_{X\rightarrow C}$ to predict concept labels $C_1,\dots, C_L$, and a \textit{target classifier} $f_{C\rightarrow Y}$ to predict the ground truth class labels $Y_1\dots,Y_K$. The details of how these models are trained are important for what follows, and we will elucidate this now. \citet{CBMs} outline three ways that these two components can be trained together: (i) \textit{Independent bottleneck} where $f_{X\rightarrow C}$ and $f_{C\rightarrow Y}$ are trained independently and the target model is trained on the ground truth concept labels; (ii) \textit{Sequential bottleneck} where $f_{X\rightarrow C}$ is trained initially, then $f_{C\rightarrow Y}$ is trained using the predicted concept labels; and (iii) \textit{Joint bottleneck} which trains both $f_{X\rightarrow C}$ and $f_{C\rightarrow Y}$ simultaneously, minimising a weighted sum of their losses and where the target model is fit using predicted concept labels.

The CLM network is performing multi-label classification, as many concepts can be true about the input data point. Thus, the outputs of the final layer of the CLM network are passed through a sigmoid activation function, to attain a vector of probabilities. Each of these probabilities correspond to the probability that their corresponding concept is true. A key distinction for this work is whether these probabilities are thresholded into \textit{hard concept labels}, or simply left as-is: \textit{soft concept labels}.

The phenomenon of information leakage occurs when soft labels are used. It was shown to occur in the joint bottleneck paradigm by \citet{margeloiu2021concept}, and later in the sequential bottleneck paradigm by \citet{mahinpei2021promises}. In the latter, they expose information leakage by conducting an experiment on the \textsc{ParityMNIST} task: given a handwritten digit from the \textsc{MNIST} dataset, determine if the digit is odd or even.
Explicitly, given an \textsc{MNIST} datapoint $(X,y)$, where $y$ is the numeric value for the number represented in the image $X$, the target label for this datapoint in the \textsc{ParityMNIST} task is $Y_0$ if $y$ is an odd number, and $Y_1$ otherwise. The concept labels $C_0,\dots, C_9$ are one-hot encodings of the original \textsc{MNIST} image labels, such that $C_i = 1$ if $i=y$ and $C_i=0$ otherwise. 
In the experiment, a CBM is trained sequentially with $f_{X\rightarrow C}$ and $f_{C\rightarrow Y}$ both being neural networks with a single hidden layer of 128 neurons. To demonstrate information leakage they consider the case where only two concepts are active, $C_3$ and $C_4$. That is, for an \textsc{MNIST} datapoint $(X,y)$ the one hot concept label is $[1,0]$ (\emph{resp.} $[0,1]$ if $y=4$), and $[0,0]$ otherwise. They report that in the case where they remove all $3$s and $4$s from the train and test sets, the CBM attains target classification accuracy of around $0.69$. The purpose of the experiment is to highlight an absurdity -- a model that can only recognise $3$ or $4$, but then sees no $3$s or $4$s should perform no better than chance on the task of predicting parity in the above setting. Trust in the model is eroded: the CBM as demonstrated does much better than chance.

This \textsc{ParityMNIST} example highlights information leakage but it is possible to understand this phenomenon in simpler terms: projecting the data $X$ on to the lower dimensional subspace spanned by the weights of the CLM model $f_{X\rightarrow C}$ preserves some structure of $X$ in the soft concept space. The structure that is preserved is often enough for the target classifier $f_{C\rightarrow Y}$ to perform well at the target prediction even when it should not be able to according to our intuition about how the concepts relate to the target labels. In the next section we will explore this notion further, with the aim of mitigating leakage but preserving the desirable quantification of predictive uncertainty provided by soft labels.
\paragraph{Why does leakage happen?}
\begin{figure}[t]
    \centering
    \includegraphics[width=1.0\textwidth]{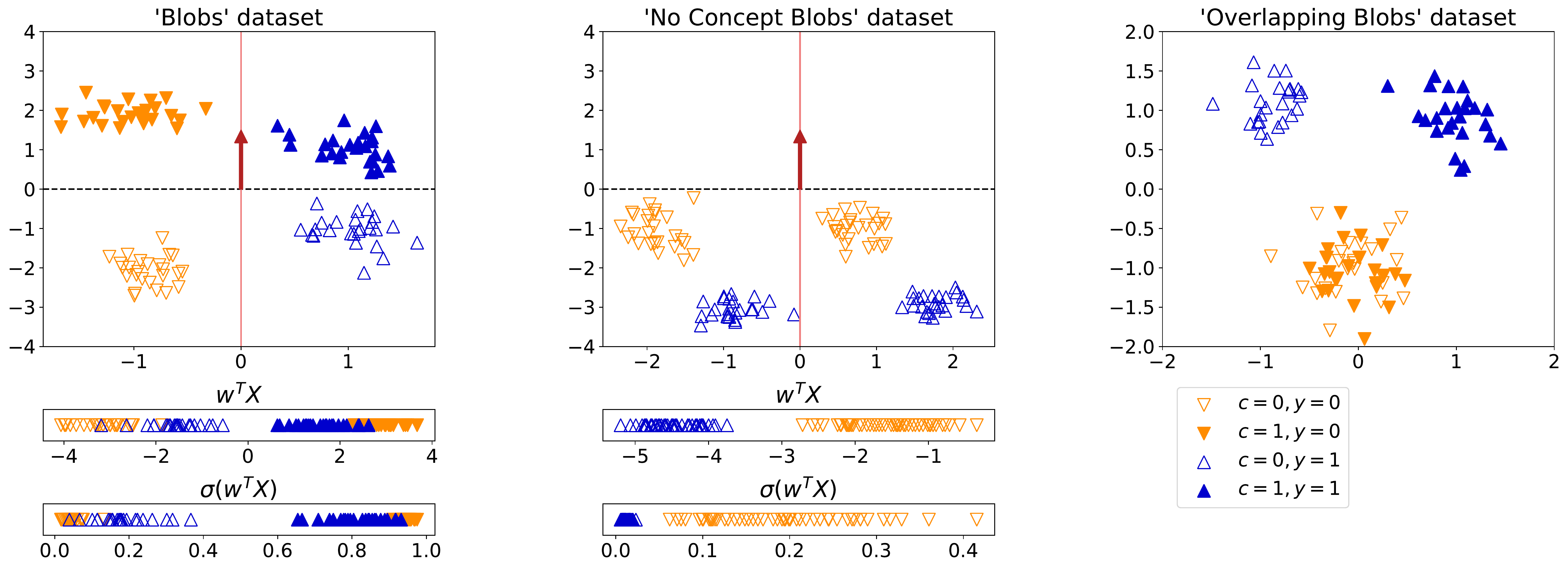}
    \caption{LEFT: leakage where uninformative concepts lead to high target accuracy, MIDDLE: leakage still happens with empty concepts, RIGHT: quantification of predictive uncertainty for concepts is necessary to detect class $Y=0$.}
    \label{fig:leakage_illustration}
\end{figure}
In Figure \ref{fig:leakage_illustration} we depict three datasets that will be helpful in understanding how leakage occurs.
Consider the dataset illustrated in the top-left plot, \textsc{Blobs}. Each datapoint is sampled from one of four normal distributions centred on $[-1.0,-2.0], [1.0,-1.0], [1.0,1.0],$ and $[-1.0,2.0]$ respectively. Each data point has a concept label $C$ and a target label $Y$. Each normal distribution corresponds to data with labels $(C=0,Y=0), (C=0,Y=1), (C=1,Y=0)$ and $(C=1,Y=1)$ respectively. The data labeling has been designed such that knowledge of the concept label $C$ provides no benefit in predicting the target label $Y$. The concept label $C$ is easily learned: the data points for concept $C=0$ are linearly separable from the data points for concept $C=1$. In the plot we show a dashed line that separates them, and its corresponding weight vector $w$ in red. In the left-middle plot of Figure \ref{fig:leakage_illustration} we show the orthogonal projection of the data onto this weight vector, $w^T X$. Finally, in the left-bottom plot we show $\sigma(w^T X)$, which correspond to the soft concept labels in the concept bottleneck paradigm.
Crucially, as evidenced in the bottom plot, the data points remain separable with respect to the target label $Y$ in the soft concept label space. 

The dataset in the top-middle plot of Figure \ref{fig:leakage_illustration} demonstrates the properties of the \textsc{ParityMNIST} example: all datapoints have $C=0$.
The CLM will learn to classify all datapoints as having $C=0$, which in the context of a single layer neural network, means it learns a weight vector corresponding to a separating line above all the datapoints.
The projection on to such a weight vector, and the application of the sigmoid to obtain soft labels, are depicted in the two lower plots.
Again, in concept space the data is separable with respect to the target label: the spatial structure of the data is preserved in the soft labels.
\paragraph{Can the leakage be mitigated?}
We understand leakage to be solely a symptom of the use of soft concept labels. Of course, this means that leakage can be completely prevented by thresholding the soft concept predictions to obtain hard labels. However, quantifying the CLM's uncertainty in its concept predictions is important.
Consider the \textsc{OverlappingBlobs} dataset in Figure \ref{fig:leakage_illustration} shown in the plot on the right hand side. In this task all data points with target label $Y=0$ are drawn from normal distributions with overlapping means: it is impossible to determine the concept label $C$ for these datapoints. A concept bottleneck model with hard concept labels will perform poorly at predicting the target label $Y$ for this dataset.
Conversely, a concept bottleneck model that can quantify its prediction uncertainty can perform well -- when the classifier is uncertain about the concept label then that datapoint is more likely to belong to class $Y=0$.

In what remains, we show how soft concept labels can be attained by applying \textit{Monte-Carlo Dropout}. Soft concept labels obtained in this way do not contain leaked information, but we also preserve the aforementioned benefits of quantifying uncertainty about concept predictions.

Dropout is a technique commonly used to reduce overfitting when training neural networks: neurons are switched on and off randomly during training. Normally, at prediction time dropout is switched off. \textit{Monte-Carlo Dropout} (MCD) is a technique for quantifying a neural network's uncertainty about its predictions based on the observation of \citet{MCD} that a neural network with dropout is an approximation of a Gaussian Process. Concretely, keeping dropout switched on at prediction time and averaging the predictions enables one to estimate the neural networks predictive uncertainty.
\section{Experiments}
The aim of this work is to mitigate the phenomenon of information leakage in CBM predictions, while maintaining the ability of the model to reason about concepts with uncertainty. To this end, we quantify the target performance accuracy of a concept bottleneck model with a variety of CLM models, and where the target classifier is the $3$-Nearest Neighbours ($3$-NN) classifier. While the standard concept bottleneck models in the literature use neural networks as the target classifier, the use of $3$-NN gives us a reliable way of determining if there is predictive power in the soft concept labels, without needing to be concerned about convergence of a target neural network. All CLM models are fully connected neural networks with a single hidden layer of 128 neurons.

\begin{table}[t]
\caption{Target label classification accuracy for different CBM configurations.}
\label{results_table}
\begin{small}
\begin{tabular}{|c|c|c|c|c|}
\cline{2-5}
\multicolumn{1}{c|}{} & \multicolumn{4}{c|}{\textbf{Mean target acc.}} \\
\hline
\textbf{Dataset} & NN+Hard+MCD & NN+Soft+MCD & NN+Hard & NN+Soft\\
\hline
\textsc{Blobs} & 0.497 & 0.837 & 0.495 & 0.856\\
\textsc{NoConceptBlobs} & 0.487 & 0.979 & 0.474 & 0.981\\
\textsc{AmbiguousBlobs} & 0.595 & 0.998 & 0.486 & 0.995\\
\textsc{OverlappingBlobs} & 0.952 & 0.993 & 0.519 & 0.997\\
\textsc{ParityMNIST} & 0.502 & 0.649 & 0.505 & 0.673\\
\textsc{ParityMNIST-NoMissing} & 0.598 & 0.627 & 0.596 & 0.642\\
\hline
\end{tabular}
\end{small}
\end{table}

In Table \ref{results_table} we list target label classification accuracy for four different concept bottleneck models characterised by their CLMs: neural network (NN) with Hard / Soft concept labels, with or without Monte-Carlo Dropout (MCD). These results are averaged over 20 repeats for the blobs based datasets and 10 repeats for the MNIST based datasets.

From the table: \textsc{Blobs} is the dataset shown in Figure \ref{fig:leakage_illustration} LEFT, which is intended to capture a dataset where concept labels are easy to predict, but are not correlated with the target labels.
Conversely, the \textsc{AmbiguousBlobs} dataset is intended to capture the setting where the hard concept labels do not correlate with the target label, but the concept labels are ambiguous and knowledge of when this ambiguity appears can be informative about the target label. This dataset is a modification of the \textsc{Blobs} dataset, where we set the means of each of the normal distributions to be
$\mu_{0,0}=[-1.0,-2.0],  \mu_{0,1}=[1.0,-0.25], \mu_{1,0}=[1.0,0.25]$, and  $\mu_{1,1}=[-1.0,2.0]$.
so that for $Y=1$, the distributions for $C=0$ and $C=1$ slightly overlap. In this way, uncertainty is informative about the target label: if the concept classifier is a little unsure about which concept label a data point should have, then that should be a sign that it is of target label $Y=1$. 

\textsc{OverlappingBlobs} is the dataset shown in Figure \ref{fig:leakage_illustration} RIGHT, where the normal distributions for label $Y=0$ sit on top of one another. The \textsc{NoConceptBlobs} dataset is the dataset depicted in Figure  \ref{fig:leakage_illustration} MIDDLE, where all concept labels are $[0,0]$ and is aimed to replicate the \textsc{ParityMNIST} example with missing $3$ and $4$ of \citet{mahinpei2021promises} in a simpler setting. Finally, we also show results for the \textsc{ParityMNIST} task with and without missing data.

\section{Discussion}
In the table above we see that a standard concept bottleneck model with soft concept labels (\textbf{NN+Soft}) attains a high classification accuracy on the \textsc{Blobs} and \textsc{NoConceptBlobs} datasets. Since the concept labels are not informative about the target labels in these datasets by design, this high accuracy must be due to leakage. Indeed, using hard concept labels (\textbf{NN+Hard}), or hard labels with MCD (\textbf{NN+Hard+MCD}) makes this accuracy go down to where it should be, near 0.5.

In the \textsc{AmbiguousBlobs} and \textsc{OverlappingBlobs} datasets, we see a boost in performance accuracy for \textbf{NN+Hard+MCD} compared to the model with no MCD (\textbf{NN+Hard}) which is what we expect: these datasets are designed so that you can get reasonable accuracy provided you are quantifying uncertainty on concept predictions. 
On the \textsc{ParityMNIST} datasets we see \textbf{NN+Hard+MCD} matches the performance of \textbf{NN+Hard} suggesting there is no gain to be had in quantifying uncertainty on this dataset.

Finally, note that for all the experimental datasets, \textbf{NN+Soft+MCD} shows higher performance accuracy than \textbf{NN+Hard+MCD}. The former technique averages the soft predictions of the CLM via MCD to attain the final prediction, while the latter averages the hard (thresholded) predictions. The fact that the former outperforms the latter suggests that information leakage also occurs in this technique via the averaging of the soft predictions. Thus, the latter should be preferred: no spurious performance boost occurs due to leakage.

The above results are vindication of our approach. Applying MCD on hard thresholded CLM predictions removes the information leakage, a fact heralded by the performance dropping on tasks where we would expect it to drop. On the other hand, our approach preserves the ability to reason about uncertainty, since performance rises again on tasks where uncertainty quantification about concepts is important.

\bibliography{refs}
\bibliographystyle{iclr2022_conference}
\subsubsection*{Disclaimer}
This paper was prepared for informational purposes by the Artificial Intelligence Research group of JPMorgan Chase \& Co.  and its affiliates (``JP Morgan''), and is not a product of the Research Department of JP Morgan. JP Morgan makes no representation and warranty whatsoever and disclaims all liability, for the completeness, accuracy or reliability of the information contained herein. This document is not intended as investment research or investment advice, or a recommendation, offer or solicitation for the purchase or sale of any security, financial instrument, financial product or service, or to be used in any way for evaluating the merits of participating in any transaction, and shall not constitute a solicitation under any jurisdiction or to any person, if such solicitation under such jurisdiction or to such person would be unlawful.

\appendix
\section{Appendix: Full results table}

\begin{tabular}{|c|c|c|c|c|c|}
\hline
Dataset & Algorithm & Mean Acc. & Std. dev. Acc. & Min. Acc. & Max. Acc. \\
\hline
\textsc{AmbiguousBlobs} & NN+Hard+MCD & 0.595 & 0.043 & 0.522 & 0.660\\
\textsc{AmbiguousBlobs} & NN+Soft+MCD & 0.998 & 0.006 & 0.981 & 1.000\\
\textsc{AmbiguousBlobs} & NN+Hard & 0.486 & 0.056 & 0.403 & 0.591\\
\textsc{AmbiguousBlobs} & NN+Soft & 0.995 & 0.005 & 0.987 & 1.000\\
\textsc{Blobs} & NN+Hard+MCD & 0.497 & 0.032 & 0.440 & 0.566\\
\textsc{Blobs} & NN+Soft+MCD & 0.837 & 0.070 & 0.717 & 0.950\\
\textsc{Blobs} & NN+Hard & 0.495 & 0.031 & 0.440 & 0.547\\
\textsc{Blobs} & NN+Soft & 0.856 & 0.084 & 0.642 & 0.962\\
\textsc{NoConceptBlobs} & NN+Hard+MCD & 0.487 & 0.033 & 0.434 & 0.547\\
\textsc{NoConceptBlobs} & NN+Soft+MCD & 0.979 & 0.015 & 0.943 & 1.000\\
\textsc{NoConceptBlobs} & NN+Hard & 0.474 & 0.033 & 0.415 & 0.522\\
\textsc{NoConceptBlobs} & NN+Soft & 0.981 & 0.020 & 0.925 & 1.000\\
\textsc{OverlappingBlobs} & NN+Hard+MCD & 0.952 & 0.086 & 0.698 & 1.000\\
\textsc{OverlappingBlobs} & NN+Soft+MCD & 0.993 & 0.005 & 0.981 & 1.000\\
\textsc{OverlappingBlobs} & NN+Hard & 0.519 & 0.045 & 0.447 & 0.597\\
\textsc{OverlappingBlobs} & NN+Soft & 0.997 & 0.003 & 0.994 & 1.000\\
\textsc{ParityMNIST} & NN+Hard+MCD & 0.502 & 0.010 & 0.487 & 0.515\\
\textsc{ParityMNIST} & NN+Soft+MCD & 0.649 & 0.006 & 0.641 & 0.663\\
\textsc{ParityMNIST} & NN+Hard & 0.505 & 0.008 & 0.487 & 0.516\\
\textsc{ParityMNIST} & NN+Soft & 0.673 & 0.018 & 0.652 & 0.713\\
\textsc{ParityMNIST-NoMissing} & NN+Hard+MCD & 0.598 & 0.003 & 0.592 & 0.602\\
\textsc{ParityMNIST-NoMissing} & NN+Soft+MCD & 0.627 & 0.010 & 0.611 & 0.649\\
\textsc{ParityMNIST-NoMissing} & NN+Hard & 0.596 & 0.009 & 0.581 & 0.607\\
\textsc{ParityMNIST-NoMissing} & NN+Soft & 0.642 & 0.016 & 0.614 & 0.661\\
\hline
\end{tabular}

\end{document}